\pdfoutput=1

\documentclass[11pt]{article}

\usepackage{EMNLP2023}

\usepackage{times}
\usepackage{amsmath}
\usepackage{latexsym}

\usepackage[T1]{fontenc}

\usepackage[utf8]{inputenc}

\usepackage{microtype}

\usepackage{inconsolata}
\usepackage{microtype}

\title{RankAug: Augmented data ranking for text classification}

\author{Tiasa Singha Roy \\
  Manipal Institute of Technology \\
  \texttt{tiasa.singharoy@gmail.com} \\\And
  Priyam Basu \\
  Manipal Institute of Technology \\
  \texttt{priyambasu16@gmail.com}}

\begin{document}
\maketitle
\begin{abstract}
Research on data generation and augmentation has been focused majorly on enhancing generation models, leaving a notable gap in the exploration and refinement of methods for evaluating synthetic data. There are several text similarity metrics within the context of generated data filtering which can impact the performance of specific Natural Language Understanding (NLU) tasks, specifically focusing on intent and sentiment classification. In this study, we propose RankAug, a text-ranking approach that detects and filters out the top augmented texts in terms of being most similar in meaning with lexical and syntactical diversity.
Through experiments conducted on multiple datasets, we demonstrate that the judicious selection of filtering techniques can yield a substantial improvement of up to 35\% in classification accuracy for under-represented classes.
\end{abstract}

\section{Introduction}
Recent advances in Large Language Models have brought along incredible progress in a wide range of NLU tasks. However, for domain specific tasks, fine-tuned models can bridge the performance gap with data \citet{wu2023exploring} but such domains are often low resource in nature and data collection can be quite difficult. This is where data augmentation techniques come into play, boosting model performance for a given supervised task by generating novel data points that are similar in characteristics to the available data. 

There have been a large number of metrics created to evaluate data augmentation which are mostly focused on the performance of generation models \citet{zhu2018texygen} \citet{kim2020collaborative} \citet{liu2020learning} \citet{sun2020mixup}. We explore various methods to evaluate and filter generated paraphrases \citet{golovneva2022task} to get high quality augmentations. Most of the prior work in this domain makes use of metrics that only take into consideration the word or embedding level similarity of the generated utterance. Popular metrics like BLEU score \citet{papineni2002bleu}, Recall-Oriented Understudy for Gisting Evaluation (ROUGE) \citet{lin2004rouge} (Lin, 2004), and Metric for Evaluation of Translation with Explicit Ordering (METEOR) \citet{banerjee2005meteor} use n-gram based comparison. This type of evaluation is limited to a one-dimensional analysis of augmentation as high-quality data provides both contextual similarity as well as lexical diversity \citet{mccarthy2009components} to the original text. To ensure this, we propose a text ranking method that outperforms other popularly used metrics to get top quality augmentations that aid in better training of models on downstream tasks. This method can be extended to any data augmentation model for evaluation and is independent of the training model as well. 

Despite a variety of approaches for augmented data evaluation, there is no golden standard \citet{bhandari2020metrics}, the real value of the generated data can be only evaluated through downstream tasks, for example by estimating how much performance improvement synthetic data can bring to the targeted supervised NLU task. In our case, we test our ranking and filtering mechanism on multiple supervised classification based scenarios for skewed datasets. It shows a consistent improvement across different experimental setups and datasets compared to the standard filtering metrics. Finally, our method \footnote{\url{https://github.com/whopriyam/Text-Augmentation-Filtering}} is also extended to a German dataset, to show that it can be applied not only to English but also to other languages.

\section{Related Works}

In recent years, data augmentation and generation techniques have gained significant attention in machine learning research. These techniques play a crucial role in enhancing the performance and robustness of various models across different domains. Augmentation techniques, in general, have been traditionally used in many downstream computer vision task \citet{kingma2013auto} uses Variational Autoencoders to encode the data examples to a latent representation and then new samples were generated from that latent space, which employs patch based augmentation. \citet{alexey2016discriminative} uses rule based image transformations to generate more data for improving performance of Convolutional Neural Networks (CNNs) on feature learning tasks. 

Text generation has been studied extensively leading to computational linguistics and diverse methods being suggested ever since. Sentence structures are very different and these diversities expand in different types of social media which makes text generation harder. Rule based techniques like word replacement using Finite State Transducers \citet{rastogi2016weighting} and synonym swap \citet{csahin2019data} have been some of the initial attempts at generating synthetic texts. Most such rule based methods suffer from a lack of sentence structure variation and loss of semantic context.

Multiple efforts have been made recently to use generative models too for text augmentation. Existing augmentation methods work at different granularity levels - characters, words, sentences, and documents. \citet{yu2018qanet} and \citet{hou2018sequence} use sequence-to-sequence generation for enhancing model performance in back translation and text transfer domains. \citet{ding2020daga} proposes a novel approach to utilize generative augmentation for fine-grained and token-level entity tagging tasks. Pre-trained masked language models (MLMs) like BERT \citet{devlin2018bert}, T5 \citet{raffel2020exploring} and AugGPT \citet{dai2023auggpt}, which internally uses ChatGPT, can be used for contextual augmentation too. Since MLMs are pre-trained on a large number of texts, contextual augmentation can usually generate meaningful new texts.

\section{Data}
In our experiments, we make use of two datasets - Airline Travel Information System (ATIS) from the Microsoft Cognitive Toolkit (CNTK) \citet{hemphill1990atis}, an intent classification dataset, Hate Speech from a white supremacist forum \citet{gibert2018hate}, a sentiment analysis dataset and Amazon Multilingual Reviews \citet{marc_reviews}, a product reviews corpus. All of these are standard datasets, ideal for setting benchmarks on classification tasks. 
\begin{itemize}
\item ATIS dataset - It consists of a set of spoken utterances in the context of airline information, classified into one of 26 intents with 127 slot labels. It is important to note that the intent distribution within the ATIS dataset exhibits a significant imbalance, with over 70\% of the data allocated to \emph{atis flight} intent, while other intents contain a notably lower number of utterances.
\item Hate Speech dataset - It consists user generated hate speech content from Stormfront, a white supremacist platform, manually annotated by human labellers. There is a high data imbalance here too, with 86\% of the texts belonging to "no hate" and 14\% belonging to "hate" sentiment.
\item Multilingual Amazon Reviews Corpus - It consists of over one million product reviews in 6 languages, ranging from 1 to 5 stars, for multilingual text classification collected between November 1, 2015 and November 1, 2019. Due to the data being sufficiently large in size, we limit our experiments to 0.5\% i.e 1000 samples, of the German reviews subset, while maintaining equal distribution across all 5 classes.
\end{itemize}

\begin{table*}
\centering
\begin{tabular}{|c|c|c|c|c|c|}
\hline
     \textbf{Dataset} & \textbf{\#} & \textbf{\#} & \textbf{\# samples} & \textbf{\# samples} & \textbf{\# samples} \\ 
      & \textbf{Classes} & \textbf{samples} & \textbf{before filtering} & \textbf{after} & \textbf{after} \\
      & & & & \textbf{RankAug-3} & \textbf{RankAug-5} \\ 
     \hline
     \textbf{ATIS Intent} & 26 & 4978 & 38358 & 9985 & 13323 \\
     \hline
     \textbf{Hate Speech}& 2 & 9666 & 16626 & 11754 & 13146 \\
     \hline
     \textbf{German Reviews} & 5 & 1000 & 11000 & 4000 & 6000 \\
     \hline
\end{tabular}
\caption{\label{citation-guide}
Benchmarking datasets
}
\end{table*}

\section{Filtering methods}
\subsection{Existing metrics}
Existing filtering metrics are efficient in assessing the quality and relevance of text content. They are of majorly two types - word based and embedding based filtering. These methods excel in their ability to capture semantic and syntactic similarities between texts, making them a preferred choice for evaluating the performance of augmented sentences. We evaluate 5 such metrics:
\begin{itemize}
\item SacreBLEU: Though is primarily used for evaluating machine translation quality, it can also be applied to filter and rank text based on translation relevance by calculating the BLEU score, which measures the similarity between the reference and the candidate sentence, by measuring the linguistic similarity and fluency of the text \citet{post2018call}. The higher the number of overlapping n-grams between candidates and source sentences, the lower the score. 
\item Levenshtein distance: In order to augmentations most similar in structure and word distribution, we use this metric. It quantifies the minimum number of single-character edits (insertions, deletions, or substitutions) required to transform one text string into another \citet{yujian2007normalized}. The lower the score, the more similar is the reference text to the source text.
\item Rouge-L: evaluates the performance of a generated text by comparing it to one or more reference texts. It considers the recall, or the ability of a generated text to capture essential information from the references, while also penalizing excessive word overlap.   \citet{lin2004rouge}.
\item Meteor: It offers a holistic evaluation by considering precision, recall, stemming, and synonymy, resulting in a more human-like assessment of text quality \citet{banerjee2005meteor}. The adaptability it provides to different languages and domains makes it a good metric to rank and filter text according to its linguistic and semantic similarity to a reference.
\item BERTScore: It leverages the pre-trained contextual embeddings from BERT and matches words in candidate and reference sentences by cosine similarity. This enables us to filter sentences that might be completely different in word measure, synonym match, sentence structure, etc but could be semantically similar in meaning \citet{bert-score}.
 \end{itemize}

\subsection{RankAug}
We propose RankAug, a ranking method that accounts for both similarity and diversity to filter high quality augmentations. 

\subsubsection{Semantic Similarity}
To measure semantic similarity we utilise BERTScore which calculates similarity scores by aligning the paraphrase $u_{i}$ and original sentence $u$ on a token-level basis. This alignment process follows a greedy approach, aiming to optimize the cosine similarity between contextualized token embeddings obtained from BERT. A higher score indicates a higher semantic relevance and we denote this as $R_{s_{i}}$ which represents the semantic rank of the $i$th paraphrase for a generated data point. 

\subsubsection{Diversity}
Diversity as an evaluation metric is often overlooked when measuring paraphrase quality. We propose self-Levenshtein ($Self$-$LD$) to compute the diversity between a generated paraphrase and both the original sentence as well as the remaining paraphrases. This is derived from self-bleu \citet{zhu2018texygen} and computes the average word-level Levenshtein distance ($LD$) i.e word error rate \citet{morris2004and} across the remaining paraphrases $u'$ and the mean is selected as the final score.
\begin{equation}
Self-LD = mean(Lev(u_{i}, u')) 
\end{equation}
This is done for every generated paraphrase with a high score indicating a higher level of diversity and the paraphrases are ranked accordingly with $R_{d{_i}}$ representing the diversity rank.

\subsubsection{Final Ranking}
After scores for both diversity and semantic similarity are generated for each paraphrase, we consider the ranking of each paraphrase based on these two criteria. We consider the harmonic mean of the generated scores to compute our final rank $R_{i}$.
\begin{equation}
R_{i} = \frac{2*R_{s_{i}}*R_{d_{i}}}{(R_{s_{i}}+R_{d_{i}})}
\end{equation}
To utilize this final rank to filter out the best paraphrases. For our experiments, we select n={3,5} values where n denotes the number of samples ranked from top i.e. top n samples.


\begin{table*}
\centering
\begin{tabular}{|c|c|ccc|}
\hline
\textbf{Filtering} & \textbf{\# augmentations  } &  & \textbf{Accuracy} & \\
\textbf{method} & \textbf{filtered per sample (n)} & \textbf{ATIS} & \textbf{Hate Speech} & \textbf{German Reviews}\\
\hline
\textbf{Baseline} &  & 98.25\% & 63\% & 50.3\%\\
\hline
\textbf{No filtering} &  & 97.35\% & 68.2\% & 48.4\%\\
\hline
\textbf{RankAug} & 5 & \textbf{99.625}\% & \textbf{74.1}\% & \textbf{54.2}\%\\
 & 3 & 98.75\% & 70.25\% & \textbf{51.4}\%\\
\hline
\textbf{Bleu} & 5 & 99.14\% & 69.8\% & 52.1\%\\
 & 3 & 98.60\% & 65\% & 45.2\%\\
\hline
\textbf{BertScore} &  5 & 99.00\% & 70.9\% & 50.1\%\\
 & 3 & 98.45\% & 68.3\% & 49.4\%\\
\hline
\textbf{Levenshtein} & 5 & 99.375\% & 70\% & 47.8\%\\
 & 3 & \textbf{99.15}\% & 69\% & 45.6\%\\
\hline
\textbf{Rouge} & 5 & 99.12\% & 72\% & 52.2\%\\
 & 3 & 98.70\% & \textbf{70.5}\% & 49.4\%\\
\hline
\textbf{Meteor} & 5 & 99.00\% & 65.7\% & 46.8\%\\
 & 3 & 98.25\% & 66.4\% & 42.8\%\\
\hline
\end{tabular}
\caption{\label{citation-guide}
Overall Accuracy for different filtering methods
}
\end{table*}

\section{Experiments}
In this section, we describe the experimental setup for benchmark tests along with the sentence generation pipeline.

\subsection{Sentence Generation}
Data sparsity is a frequent problem for several NLU tasks as collecting the necessary quantities of high-quality labeled data for model training is frequently a challenging and expensive task, along with the risks of the generative model becoming too large \citet{bender2021dangers}. We undertake the task to produce artificial data that can be utilized to enhance NLU model training. We use the original training data from the corpora as a source to the data generation model.

For augmenting the English sentences, we leveraged Google's transformer-based Pegasus model \citet{zhang2020pegasus} for text augmentation. Pegasus internally uses self-supervised gap sentence generation for better abstraction performance by masking important tokens and applying ROGUE-n selection. It was pre-trained on the Colossal Common Crawl C4 \citet{dodge2021documenting} dataset. We used a pretrained Pegasus model fine-tuned on Google Paws \citet{pawsx2019emnlp}, which is a paraphrasing dataset as a one-to-one sentence generator. By limiting the paraphrase token length limit, abstracting from a short sentence, the model paraphrases the text to a semantically similar sentence. 

For German text augmentation, we use a pivot-based back translation pipeline \citet{cai2021revisiting}. In this process, the input texts are first translated to pivots and then paraphrases are generated. The German texts are first converted to English, which are then used as a pivot to generate the required paraphrases.

\subsection{Downstream Task}
For the purpose of evaluating the quality of generated text, we chose a text classification task. We use BERT-base embeddings for ATIS and Hate speech classification and, Bert-base multilingual embeddings for classifying the German Amazon reviews. Both models are trained for 4 epochs along with a batch size of 16 and a learning rate of 2e-5. Adding the corpus augmented with paraphrase improves the performance, which shows that it helps training even when fine-tuning the pre-trained language model. We average out the results from 3 runs for each scenario. Performance changes through data augmentation are significant, especially when the baseline accuracy is less, which is evident as shown in Table 5.

\section{Results}
For the ATIS and Hate Speech datasets, which are imbalanced datasets, we generate paraphrases for each underrepresented data point. This excludes the \emph{atis flight} and \emph{not hate} classes for ATIS and hate speech respectively. This gives us around 38358 for ATIS and 16626 samples in total for hate speech. However, models trained using the original data and every generated paraphrase result in a decrease in performance, highlighting the importance of using quality data points.

We then apply different filtering methods on the generated data. All our filtering methods showcase a consistent improvement on both datasets and outperform the baseline. As evident in Table 5, our method shows consistent improvement on the ATIS dataset and four other filtering methods show better results across both datasets as compared to the baseline. This shows that by utilizing less than half of the generated data we can outperform the baseline. 

\begin{table}[h!]
    \label{tab:table2}
    \centering
    \begin{tabular}{|c|c|c|c|c|}
    \hline
         \textbf{Intent} & \textbf{Train} & \textbf{Test} & \textbf{Base-} & \textbf{Rank} \\
         \textbf{} & \textbf{} & \textbf{} & \textbf{line} & \textbf{Aug-5} \\
         \hline
         \textbf{airfare} & 385 & 48 & 95.83\% & 97.92\% \\
         \hline
         \textbf{service} & 230 & 36 & 94.44\% & 100\%\\
         \hline
         \textbf{flight} & 3309 & 632 & 98.25\% & 99.47\%\\
         \hline
         \textbf{abbre-} & 130 & 33 & 96.96\% & 100\%\\
         \textbf{viation} &  &  &  & \\
         \hline
         \textbf{airline} & 139 & 38 & 94.74\% & 100\%\\
         \hline
    \end{tabular}
    \caption{RankAug-5 performance on ATIS dataset on top 5 intents}
\end{table}

\begin{table}[h!]
    \label{tab:table2}
    \centering
    \begin{tabular}{|c|c|c|c|c|}
    \hline
         \textbf{Senti-} & \textbf{Train} & \textbf{Test} & \textbf{Base-} & \textbf{Rank} \\
         \textbf{ment} & \textbf{} & \textbf{} & \textbf{line} & \textbf{Aug-5} \\
         \hline
         \textbf{Hate} & 696 & 500 & 19.02\% & 54.78\% \\
         \hline
         \textbf{Not hate} & 8970 & 500 & 97.60\% & 100\%\\
         \hline
    \end{tabular}
    \caption{RankAug-5 performance on Hate speech dataset per sentiment}
\end{table}

\begin{table}[h!]
    \label{tab:table2}
    \centering
    \begin{tabular}{|c|c|c|c|c|}
    \hline
         \textbf{Rating} & \textbf{Train} & \textbf{Test} & \textbf{Base-} & \textbf{Rank} \\
         \textbf{} & \textbf{} & \textbf{} & \textbf{line} & \textbf{Aug-5} \\
         \hline
         \textbf{1 star} & 200 & 100 & 52\% & 61\% \\
         \hline
         \textbf{2 star} & 200 & 104 & 41.34\% & 54.81\%\\
         \hline
         \textbf{3 star} & 200 & 105 & 39.12\% & 49.91\%\\
         \hline
         \textbf{4 star} & 200 & 99 & 29.29\% & 34.34\%\\
         \hline
         \textbf{5 star} & 200 & 92 & 76.04\% & 75.89\%\\
         \hline
    \end{tabular}
    \caption{RankAug-5 performance on German Reviews dataset per rating}
\end{table}

Comparing against the different metrics, our method gives an increased performance for both datasets in n=5 filtered setting. For n=3, our filtering method comes close to matching the top performer. Across all 3 datasets, our method shows the best overall performance showing an increase compared to both the baseline performance and the no-filtering setting. In the case of ATIS and German review datasets, it can also be noted that no filtering augmentation setting actually reduces the performance when compared to the baseline. This essentially infers the fact that just adding augmentations does not necessarily improve performance but the generated text being of good quality is what yields good results on downstream tasks. 

In Tables 2 and 3, we can see the performance improvement across different classes. We also extend our work on a balanced but a low resource German dataset. Our filtering method outperforms all other methods for both settings indicating that RankAug is adaptable to other languages as well.\\

\section{Conclusion}

When working with low resource and unbalanced datasets, data augmentation can significantly improve performance. However, it is crucial to have quality augmented data. We explored and evaluated a number of popular evaluation metrics for augmented data filtering and proposed our own method for ranking and filtering quality paraphrases. Our method, along with similarity also accounts for paraphrase diversity and achieves the best overall performance across multiple datasets while utilizing nearly half the total augmented data. Along with this, we also observe a consistent increase in performance of the underrepresented classes of the datasets explored achieving up to 35\% increase in accuracy. We show that our approach can be extended to other languages as well as other varied domains to improve downstream performance and as a future work, we aim to benchmark these methods on more downstream tasks.

\section*{Limitations}
While we achieve improvements in the datasets selected,wh the methods required to generate paraphrases can be very resource heavy. Along with this, BERTScore also requires GPU resources and is time consuming to use especially when a large amount of augmented data is present. For testing, we only consider downstream classification tasks which limit our evaluation as other tasks can have different requirements that our method is not able to encompass and should be explored. 

\bibliography{anthology,custom}

\begin{thebibliography}{33}
\expandafter\ifx\csname natexlab\endcsname\relax\def\natexlab#1{#1}\fi

\bibitem[{Alexey et~al.(2016)Alexey, Fischer, Tobias, Springenberg, and
  Brox}]{alexey2016discriminative}
Dosovitskiy Alexey, Philipp Fischer, Jost Tobias, Martin~Riedmiller
  Springenberg, and Thomas Brox. 2016.
\newblock Discriminative unsupervised feature learning with exemplar
  convolutional neural networks.
\newblock \emph{IEEE TPAMI}, 38(9):1734--1747.

\bibitem[{Banerjee and Lavie(2005)}]{banerjee2005meteor}
Satanjeev Banerjee and Alon Lavie. 2005.
\newblock Meteor: An automatic metric for mt evaluation with improved
  correlation with human judgments.
\newblock In \emph{Proceedings of the acl workshop on intrinsic and extrinsic
  evaluation measures for machine translation and/or summarization}, pages
  65--72.

\bibitem[{Bender et~al.(2021)Bender, Gebru, McMillan-Major, and
  Shmitchell}]{bender2021dangers}
Emily~M Bender, Timnit Gebru, Angelina McMillan-Major, and Shmargaret
  Shmitchell. 2021.
\newblock On the dangers of stochastic parrots: Can language models be too big?
\newblock In \emph{Proceedings of the 2021 ACM conference on fairness,
  accountability, and transparency}, pages 610--623.

\bibitem[{Bhandari et~al.(2020)Bhandari, Gour, Ashfaq, and
  Liu}]{bhandari2020metrics}
Manik Bhandari, Pranav Gour, Atabak Ashfaq, and Pengfei Liu. 2020.
\newblock Metrics also disagree in the low scoring range: Revisiting
  summarization evaluation metrics.
\newblock \emph{Proceedings of the 28th International Conference on
  Computational Linguistics}, pages 5702--5711.

\bibitem[{Cai et~al.(2021)Cai, Cao, and Wan}]{cai2021revisiting}
Yitao Cai, Yue Cao, and Xiaojun Wan. 2021.
\newblock Revisiting pivot-based paraphrase generation: Language is not the
  only optional pivot.
\newblock In \emph{Proceedings of the 2021 conference on empirical methods in
  natural language processing}, pages 4255--4268.

\bibitem[{Dai et~al.(2023)Dai, Liu, Liao, Huang, Cao, Wu, Zhao, Xu, Liu, Liu
  et~al.}]{dai2023auggpt}
Haixing Dai, Zhengliang Liu, Wenxiong Liao, Xiaoke Huang, Yihan Cao, Zihao Wu,
  Lin Zhao, Shaochen Xu, Wei Liu, Ninghao Liu, et~al. 2023.
\newblock Auggpt: Leveraging chatgpt for text data augmentation.
\newblock \emph{arXiv preprint arXiv:2302.13007}.

\bibitem[{de~Gibert et~al.(2018)de~Gibert, Perez, Garc{\'\i}a-Pablos, and
  Cuadros}]{gibert2018hate}
Ona de~Gibert, Naiara Perez, Aitor Garc{\'\i}a-Pablos, and Montse Cuadros.
  2018.
\newblock \href {https://doi.org/10.18653/v1/W18-5102} {{Hate Speech Dataset
  from a White Supremacy Forum}}.
\newblock In \emph{Proceedings of the 2nd Workshop on Abusive Language Online
  ({ALW}2)}, pages 11--20, Brussels, Belgium. Association for Computational
  Linguistics.

\bibitem[{Devlin et~al.(2018)Devlin, Chang, Lee, and
  Toutanova}]{devlin2018bert}
Jacob Devlin, Ming-Wei Chang, Kenton Lee, and Kristina Toutanova. 2018.
\newblock Bert: Pre-training of deep bidirectional transformers for language
  understanding.
\newblock \emph{Proceedings of the 2019 Conference of the North American
  Chapter of the Association for Computational Linguistics: Human Language
  Technologies, Volume 1 (Long and Short Papers)}, pages 4171--4186.

\bibitem[{Ding et~al.(2020)Ding, Liu, Bing, Kruengkrai, Nguyen, Joty, Si, and
  Miao}]{ding2020daga}
Bosheng Ding, Linlin Liu, Lidong Bing, Canasai Kruengkrai, Thien~Hai Nguyen,
  Shafiq Joty, Luo Si, and Chunyan Miao. 2020.
\newblock Daga: Data augmentation with a generation approach for low-resource
  tagging tasks.
\newblock \emph{Proceedings of the 2020 Conference on Empirical Methods in
  Natural Language Processing}, pages 6045--6057.

\bibitem[{Dodge et~al.(2021)Dodge, Sap, Marasovi{\'c}, Agnew, Ilharco,
  Groeneveld, Mitchell, and Gardner}]{dodge2021documenting}
Jesse Dodge, Maarten Sap, Ana Marasovi{\'c}, William Agnew, Gabriel Ilharco,
  Dirk Groeneveld, Margaret Mitchell, and Matt Gardner. 2021.
\newblock Documenting large webtext corpora: A case study on the colossal clean
  crawled corpus.
\newblock \emph{Proceedings of the 2021 Conference on Empirical Methods in
  Natural Language Processing}, pages 1286--1305.

\bibitem[{Golovneva et~al.(2022)Golovneva, Wei, Abboud, Peris, Tan, and
  Yu}]{golovneva2022task}
Olga Golovneva, Pan Wei, Khadige Abboud, Charith Peris, Lizhen Tan, and Haiyang
  Yu. 2022.
\newblock Task-driven augmented data evaluation.

\bibitem[{Hemphill et~al.(1990)Hemphill, Godfrey, and
  Doddington}]{hemphill1990atis}
Charles~T Hemphill, John~J Godfrey, and George~R Doddington. 1990.
\newblock The atis spoken language systems pilot corpus.
\newblock In \emph{Speech and Natural Language: Proceedings of a Workshop Held
  at Hidden Valley, Pennsylvania, June 24-27, 1990}.

\bibitem[{Hou et~al.(2018)Hou, Liu, Che, and Liu}]{hou2018sequence}
Yutai Hou, Yijia Liu, Wanxiang Che, and Ting Liu. 2018.
\newblock Sequence-to-sequence data augmentation for dialogue language
  understanding.
\newblock \emph{Proceedings of the 27th International Conference on
  Computational Linguistics}, pages 1234--1245.

\bibitem[{Keung et~al.(2020)Keung, Lu, Szarvas, and Smith}]{marc_reviews}
Phillip Keung, Yichao Lu, György Szarvas, and Noah~A. Smith. 2020.
\newblock The multilingual amazon reviews corpus.
\newblock In \emph{Proceedings of the 2020 Conference on Empirical Methods in
  Natural Language Processing}.

\bibitem[{Kim et~al.(2020)Kim, Won, Yoon, and Jung}]{kim2020collaborative}
Yanghoon Kim, Seungpil Won, Seunghyun Yoon, and Kyomin Jung. 2020.
\newblock Collaborative training of gans in continuous and discrete spaces for
  text generation.
\newblock \emph{IEEE Access}, 8:226515--226523.

\bibitem[{Kingma and Welling(2013)}]{kingma2013auto}
Diederik~P Kingma and Max Welling. 2013.
\newblock Auto-encoding variational bayes.
\newblock \emph{arXiv preprint arXiv:1312.6114}.

\bibitem[{Lin(2004)}]{lin2004rouge}
Chin-Yew Lin. 2004.
\newblock Rouge: A package for automatic evaluation of summaries.
\newblock In \emph{Text summarization branches out}, pages 74--81.

\bibitem[{Liu et~al.(2020)Liu, Yang, Xiong, Zhang, Meng, Hu, Xu, and
  Chen}]{liu2020learning}
Mingtong Liu, Erguang Yang, Deyi Xiong, Yujie Zhang, Yao Meng, Changjian Hu,
  Jinan Xu, and Yufeng Chen. 2020.
\newblock A learning-exploring method to generate diverse paraphrases with
  multi-objective deep reinforcement learning.
\newblock In \emph{Proceedings of the 28th International Conference on
  Computational Linguistics}, pages 2310--2321.

\bibitem[{McCarthy et~al.(2009)McCarthy, Guess, and
  McNamara}]{mccarthy2009components}
Philip~M McCarthy, Rebekah~H Guess, and Danielle~S McNamara. 2009.
\newblock The components of paraphrase evaluations.
\newblock \emph{Behavior Research Methods}, 41(3):682--690.

\bibitem[{Morris et~al.(2004)Morris, Maier, and Green}]{morris2004and}
Andrew~Cameron Morris, Viktoria Maier, and Phil Green. 2004.
\newblock From wer and ril to mer and wil: improved evaluation measures for
  connected speech recognition.
\newblock In \emph{Eighth International Conference on Spoken Language
  Processing}.

\bibitem[{Papineni et~al.(2002)Papineni, Roukos, Ward, and
  Zhu}]{papineni2002bleu}
Kishore Papineni, Salim Roukos, Todd Ward, and Wei-Jing Zhu. 2002.
\newblock Bleu: a method for automatic evaluation of machine translation.
\newblock In \emph{Proceedings of the 40th annual meeting of the Association
  for Computational Linguistics}, pages 311--318.

\bibitem[{Post(2018)}]{post2018call}
Matt Post. 2018.
\newblock A call for clarity in reporting bleu scores.
\newblock \emph{Proceedings of the Third Conference on Machine Translation
  (WMT), Volume 1: Research Papers}, pages 186--191.

\bibitem[{Raffel et~al.(2020)Raffel, Shazeer, Roberts, Lee, Narang, Matena,
  Zhou, Li, and Liu}]{raffel2020exploring}
Colin Raffel, Noam Shazeer, Adam Roberts, Katherine Lee, Sharan Narang, Michael
  Matena, Yanqi Zhou, Wei Li, and Peter~J Liu. 2020.
\newblock Exploring the limits of transfer learning with a unified text-to-text
  transformer.
\newblock \emph{The Journal of Machine Learning Research}, 21(1):5485--5551.

\bibitem[{Rastogi et~al.(2016)Rastogi, Cotterell, and
  Eisner}]{rastogi2016weighting}
Pushpendre Rastogi, Ryan Cotterell, and Jason Eisner. 2016.
\newblock Weighting finite-state transductions with neural context.
\newblock In \emph{Proceedings of the 2016 conference of the North American
  chapter of the Association for Computational Linguistics: human language
  technologies}, pages 623--633.

\bibitem[{{\c{S}}ahin and Steedman(2019)}]{csahin2019data}
G{\"o}zde~G{\"u}l {\c{S}}ahin and Mark Steedman. 2019.
\newblock Data augmentation via dependency tree morphing for low-resource
  languages.
\newblock \emph{Proceedings of the 2018 Conference on Empirical Methods in
  Natural Language Processing}, pages 5004--5009.

\bibitem[{Sun et~al.(2020)Sun, Xia, Yin, Liang, Yu, and He}]{sun2020mixup}
Lichao Sun, Congying Xia, Wenpeng Yin, Tingting Liang, Philip~S Yu, and Lifang
  He. 2020.
\newblock Mixup-transformer: dynamic data augmentation for nlp tasks.
\newblock \emph{Proceedings of the 28th International Conference on
  Computational Linguistics}, pages 3436--3440.

\bibitem[{Wu et~al.(2023)Wu, Zhang, Cao, Yu, Dai, Ma, Liu, Zhao, Li, Liu
  et~al.}]{wu2023exploring}
Zihao Wu, Lu~Zhang, Chao Cao, Xiaowei Yu, Haixing Dai, Chong Ma, Zhengliang
  Liu, Lin Zhao, Gang Li, Wei Liu, et~al. 2023.
\newblock Exploring the trade-offs: Unified large language models vs local
  fine-tuned models for highly-specific radiology nli task.

\bibitem[{Yang et~al.(2019)Yang, Zhang, Tar, and Baldridge}]{pawsx2019emnlp}
Yinfei Yang, Yuan Zhang, Chris Tar, and Jason Baldridge. 2019.
\newblock {PAWS-X: A Cross-lingual Adversarial Dataset for Paraphrase
  Identification}.
\newblock In \emph{Proc. of EMNLP}.

\bibitem[{Yu et~al.(2018)Yu, Dohan, Luong, Zhao, Chen, Norouzi, and
  Le}]{yu2018qanet}
Adams~Wei Yu, David Dohan, Minh-Thang Luong, Rui Zhao, Kai Chen, Mohammad
  Norouzi, and Quoc~V Le. 2018.
\newblock Qanet: Combining local convolution with global self-attention for
  reading comprehension.
\newblock \emph{ICLR}.

\bibitem[{Yujian and Bo(2007)}]{yujian2007normalized}
Li~Yujian and Liu Bo. 2007.
\newblock A normalized levenshtein distance metric.
\newblock \emph{IEEE transactions on pattern analysis and machine
  intelligence}, 29(6):1091--1095.

\bibitem[{Zhang et~al.(2020)Zhang, Zhao, Saleh, and Liu}]{zhang2020pegasus}
Jingqing Zhang, Yao Zhao, Mohammad Saleh, and Peter Liu. 2020.
\newblock Pegasus: Pre-training with extracted gap-sentences for abstractive
  summarization.
\newblock In \emph{International Conference on Machine Learning}, pages
  11328--11339. PMLR.

\bibitem[{Zhang* et~al.(2020)Zhang*, Kishore*, Wu*, Weinberger, and
  Artzi}]{bert-score}
Tianyi Zhang*, Varsha Kishore*, Felix Wu*, Kilian~Q. Weinberger, and Yoav
  Artzi. 2020.
\newblock \href {https://openreview.net/forum?id=SkeHuCVFDr} {Bertscore:
  Evaluating text generation with bert}.
\newblock In \emph{International Conference on Learning Representations}.

\bibitem[{Zhu et~al.(2018)Zhu, Lu, Zheng, Guo, Zhang, Wang, and
  Yu}]{zhu2018texygen}
Yaoming Zhu, Sidi Lu, Lei Zheng, Jiaxian Guo, Weinan Zhang, Jun Wang, and Yong
  Yu. 2018.
\newblock Texygen: A benchmarking platform for text generation models.
\newblock In \emph{The 41st international ACM SIGIR conference on research \&
  development in information retrieval}, pages 1097--1100.

\end{thebibliography}
\bibliographystyle{acl_natbib}




\end{document}